\documentclass{article}

    \usepackage[nonatbib,preprint]{neurips_2025}

\usepackage[utf8]{inputenc} 
\usepackage[T1]{fontenc}    
\usepackage{hyperref}       
\usepackage{url}            
\usepackage{booktabs}       
\usepackage{amsfonts}       
\usepackage{nicefrac}       
\usepackage{microtype}      
\usepackage{xcolor}         

\usepackage{amsmath}
\usepackage{graphicx}
\usepackage{multirow}

\title{Extracting Visual Facts from
Intermediate Layers for Mitigating Hallucinations in
Multimodal Large Language Models}

\author{%
  Haoran Zhou \\
  Southeast University\\
  \texttt{haoranzhou0@gmail.com}\\
  \And
  Zihan Zhang \\
  Southeast University\\
  \texttt{220242424@seu.edu.cn} \\
  \And
  Hao Chen\thanks{Corresponding Author.} \\
  Southeast University\\
  \texttt{haochen303@seu.edu.cn}\\
}

\begin{document}

\maketitle

\begin{abstract}
Multimodal Large Language Models (MLLMs) have made significant strides by combining visual recognition and language understanding to generate content that is both coherent and contextually accurate. However, MLLMs continue to struggle with object hallucinations, where models produce seemingly plausible but factually incorrect outputs, including objects that do not exist in the image. Recent work has revealed that the prior knowledge in MLLMs significantly suppresses visual information in deep layers, causing hallucinatory outputs. However, how these priors suppress visual information at the intermediate layer stage in MLLMs remains unclear. We observe that visual factual knowledge and the differences between intermediate-layer prior/original probability distributions show similar evolutionary trends in intermediate layers. Motivated by this, we introduce Decoding by \textbf{E}xtracting \textbf{V}isual F\textbf{A}cts (\textbf{EVA}), a simple, training-free method that dynamically selects intermediate layers with the most significant visual factual information. By contrasting the output distributions of the selected layer derived from the original input and pure-text input, EVA extracts visual factual knowledge and proportionally incorporates it into the final layer to correct the output logits. Importantly, EVA is model-agnostic, seamlessly integrates with various classic decoding strategies, and is applicable across different MLLMs. We validate EVA on widely-used benchmarks, and the results show that it significantly reduces hallucination rates compared to baseline methods, underscoring its effectiveness in mitigating hallucinations.
\end{abstract}

\section{Introduction}
\label{sec:1}
Recently, the accelerated advancement of Multimodal Large Language Models (MLLMs) has emerged as a promising trajectory toward realizing Artificial General Intelligence (AGI)~\cite{wang2024qwen2,yao2024minicpm,lu2024deepseek,team2024chameleon,achiam2023gpt,liu2024improved,chern2024anole}. Nevertheless, the practical development of MLLMs is significantly constrained by the persistent challenge of hallucination, a phenomenon wherein models generate assertions about non-existent visual content while omitting critical descriptions of actually present objects, thereby propagating self-deceptive outputs~\cite{bai2024hallucination,liu2024survey,li2023evaluating,liu2023mitigating,rawte2023survey}. This issue poses profound risks in high-consequence domains such as medical imaging analysis~\cite{chen2024huatuogpt,hu2023advancing,wang2023chatcad}, autonomous driving systems~\cite{cui2024survey,wang2023drivemlm}, and human-computer interaction~\cite{brie2023evaluating}, where erroneous outputs could lead to irreversible outcomes.

The underlying mechanisms of hallucinations in MLLMs are inherently complex. Unlike prior investigations focused on Large Language Models (LLMs)~\cite{chuang2023dola,chen2024context,orgad2024llms,chen2024llama,lu2024insights,wang2024knowledge}, contemporary research~\cite{wang2024mllm, liang2024unleashing} has revealed that MLLMs possess the capacity to encode visual information. Empirical evidence has demonstrated that, to a significant extent, the prior knowledge embedded within MLLMs, particularly the textual priors of the LLM serving as the decoder, exerts suppressive effects on visual features in deeper network layers, thereby inducing hallucinatory outputs. Nevertheless, the precise mechanisms through which such priors suppress visual information at the intermediate layer stage in MLLMs remain largely underexplored.

To fill this gap, we leverage the early-exit method~\cite{teerapittayanon2016branchynet, elbayad2019depth, schuster2022confident} to derive next-token probability distributions from intermediate layers for both the original multimodal input (containing visual-textual features) and the pure text input (lacking visual components). We then employ Jensen-Shannon (JS) divergence to quantify the discrepancy between these two distributions for each layer, operationalizing this metric as a measure of how prior knowledge suppresses visual information. Empirical observations reveal that the trend of JS divergence across shallow to deep layers correlates with the probability dynamics of ground-truth tokens(as shown in Figure~\ref{fig:2} and Section~\ref{sec:3.2}). This suggests that \textbf{the intermediate layer exhibiting the maximum JS divergence harbors richer visual factual knowledge.} Building on this insight, we introduce the Decoding by \textbf{E}xtracting \textbf{V}isual F\textbf{A}ct (EVA) method, which identifies visual factual information by contrasting outputs from the selected intermediate layer of the multimodal input with those of the text-only input, thereby correcting hallucinatory outputs during generation. Notably, EVA is training-free and compatible with mainstream decoding strategies, including greedy search, nucleus sampling, and beam search, enabling seamless integration into any MLLM architecture for hallucination mitigation.

\textbf{Contributions.} Our contributions primarily involve investigating how prior knowledge in MLLMs suppresses visual factual knowledge at the intermediate layer stage. We uncovered that visual factual knowledge shares a similar evolutionary trend with the divergence between original and prior probability distributions in intermediate layers. Based on this finding, we propose EVA, a decoding method that refines final outputs by extracting visual factual knowledge from intermediate layers. EVA is seamlessly integrated with four representative MLLMs (InstructBLIP, MiniGPT-4, LLaVA-1.5, and Qwen-VL) using three classic decoding strategies: greedy search, nucleus sampling, and beam search. Experimental results demonstrate that EVA significantly outperforms baselines on visual question answering (VQA) benchmarks (POPE, MME) and achieves superior performance of image captioning datasets.

\begin{figure}[htbp]
\begin{center}
   \includegraphics[width=\linewidth]{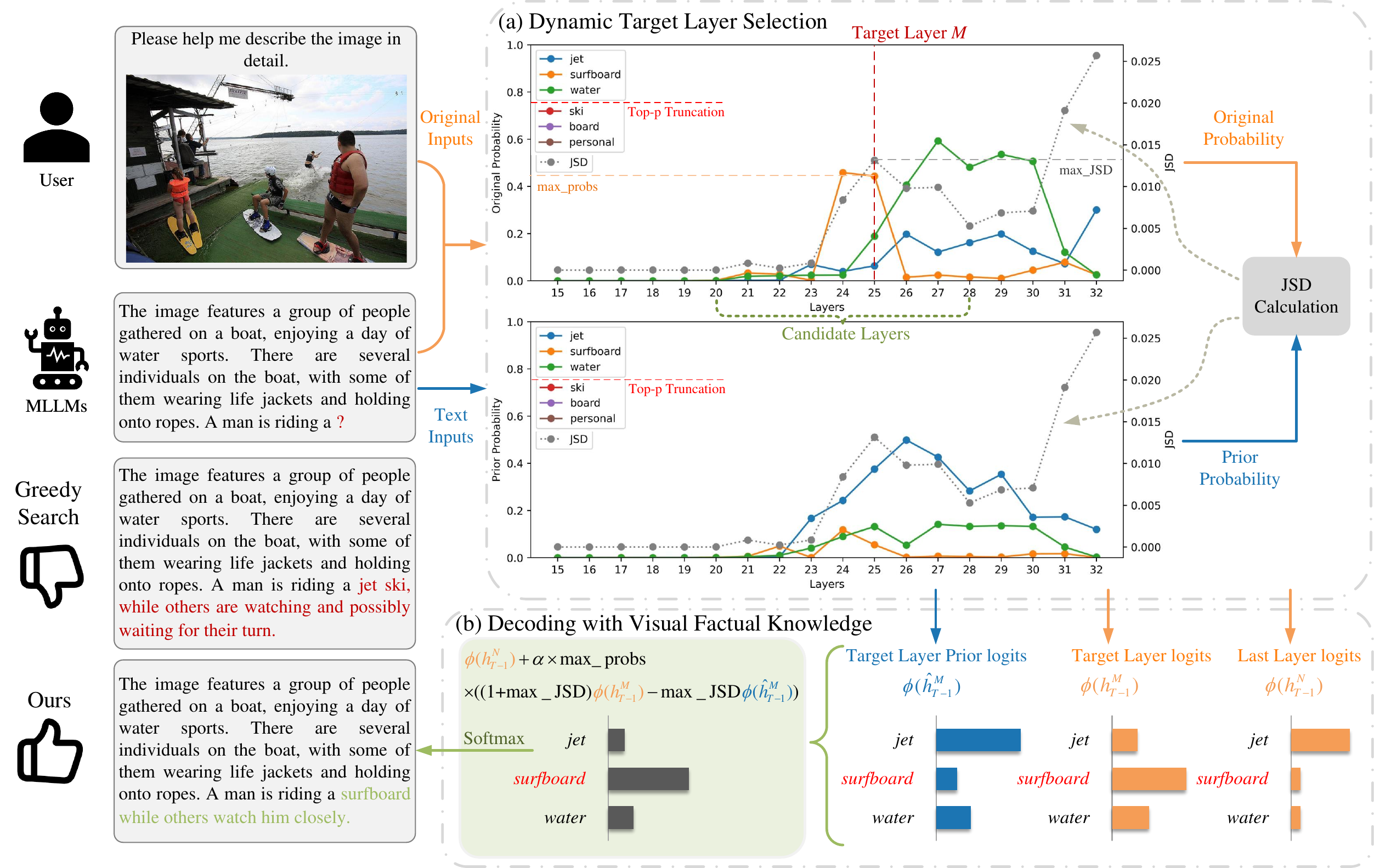}
\end{center}
   \caption{
   Framework of EVA. (a) Dynamically select an optimal target layer from intermediate layers. (b) Extract visual facts and correct the knowledge in the final layer with visual factual knowledge.}
\label{fig:1}
\end{figure}

\section{Related works}
\label{sec:3}
\subsection{Causes of Hallucination in MLLMs}
Hallucinations are frequently observed in MLLMs when processing visual and textual information. The underlying causes are multifaceted, involving complex interactions and mechanisms across different levels~\cite{jiang2024hallucination}. The inherent differences in feature representation between visual and textual modalities, along with the difficulty of accurate alignment, lead to the attenuation of visual information during cross-modal fusion. This limits the model's ability to effectively utilize visual signals internally~\cite{bai2024hallucination,lee2024delve}. Additionally, inaccurate image-text matching and biases in training datasets, such as numerous similar images paired with homogenized descriptions, reinforce the model’s reliance on statistical priors~\cite{leng2024mitigating,jiang2024hallucination}. This causes the model to frequently produce "common" but inaccurate answers that do not faithfully correspond to the inputs. Factors such as insufficient cross-modal representation alignment, imbalanced attention allocation, the overshadowing effect of knowledge priors, and hierarchical information attenuation~\cite{bai2024hallucination,wang2025mirage} accumulate to make hallucination an unavoidable core challenge in current multimodal models. Moreover, during both model training and inference, strong language priors often suppress or override the perception of visual content, causing the model to favor high-probability linguistic content while ignoring or fabricating objects and attributes absent from the image ~\cite{leng2024mitigating,lee2024delve}. Some works~\cite{wang2024mllm, liang2024unleashing} have partially validated this point, but the process by which prior knowledge suppresses visual knowledge at the intermediate layer remains unclear.

\subsection{Mitigating Hallucination for MLLMs}
 Recent research on hallucination mitigation in MLLMs has primarily focused on two main avenues: representation optimization during training and decoding strategies. Hallucination Augmented Contrastive Learning~\cite{jiang2024hallucination} introduces hallucinated text as hard negative samples in contrastive learning, strengthening the consistency between non-hallucinatory text and visual representations and improving the model's ability to distinguish between real and fabricated content. HalluciDoctor~\cite{yu2024hallucidoctor} tackles hallucination directly at the data source by automatically detecting and removing harmful hallucinatory samples from training data, and balances data distribution through counterfactual sample augmentation, fundamentally enhancing the model’s resistance to hallucination. Additionally, unsupervised decoding and semantic feedback mechanisms such as HELPD~\cite{zou2024look} and MemVR~\cite{yuan2024helpd} have been proposed, providing diversified technical solutions for MLLM hallucination mitigation. Collectively, these approaches have significantly enhanced the trustworthiness and robustness of multimodal large models in practical applications. OPERA~\cite{huang2024opera} method introduces an “over-trust penalty” in the decoding combined with a retrospection-allocation mechanism, which dynamically adjusts attention distribution and generation weights.This reduces the model’s over-reliance on a limited set of tokens and better leverages image information without requiring additional training resources. Visual Contrastive Decoding (VCD)~\cite{leng2024mitigating} is a training-free approach that suppresses hallucinations by contrasting output distributions from original and perturbed visual inputs, thereby reducing dependence on statistical biases and language priors. DeCo~\cite{wang2024mllm} adaptively selects the appropriate preceding layers and proportionally integrates knowledge into the final layer to adjust the output logits. Compared with existing decoding methods, our approach can extract visual factual knowledge and integrate it into the final output to mitigate hallucinations.

\section{Method}

\subsection{Preliminary}

\textbf{MLLM generation.} MLLMs generally concatenate visual tokens,  processed by a visual encoder and projection layer, with embedded textual tokens before feeding them into an autoregressive language model. We denote the visual token sequence as $\mathbf{X}^V = \{x_{v1}, x_{v2}, \ldots, x_{vP}\}$ and the textual token sequence as $\mathbf{X}^C = \{x_{c1}, x_{c2}, \ldots, x_{cQ}\}$, where $P$ and $Q$ represent the lengths of the visual and textual token sequences, respectively. The final input is thus $\mathbf{X} = \text{concat}(\mathbf{X}^V, \mathbf{X}^C)$. This concatenated input $\mathbf{X}$ is then passed through an MLLM comprising $N$ stacked transformer layers. The intermediate variables generated by the $i$-th layer, referred to as hidden states, are denoted as $\mathbf{h}^i = \{h_0^i, h_1^i, \ldots, h_{T-1}^i\}$, where $T = P + Q$ represents the total sequence length.  

During the generation phase, the hidden state at the final position of the last layer is mapped to the vocabulary dimension via an affine layer $\phi(\cdot)$ to predict the probability distribution over next tokens. Formally, this process is expressed as: 
\begin{equation}
    p\left(x_T\mid x_{<T}\right)=\mathrm{softmax}\left(\phi\left(h_{T-1}^N\right)\right)_{x_T},x_T\in\mathbf{\mathcal{V}}
\end{equation}
where we use $x_{<T}$ to simplify the sequence $\{{{x}_i}\}_{i=0}^{T-1}$ and $\mathcal{V}$ refers to the whole vocabulary set.

\subsection{The Intermediate Layers Contain Visual Factual Knowledge}
\label{sec:3.2}

\begin{figure}[htbp]
\begin{center}
   \includegraphics[width=\linewidth]{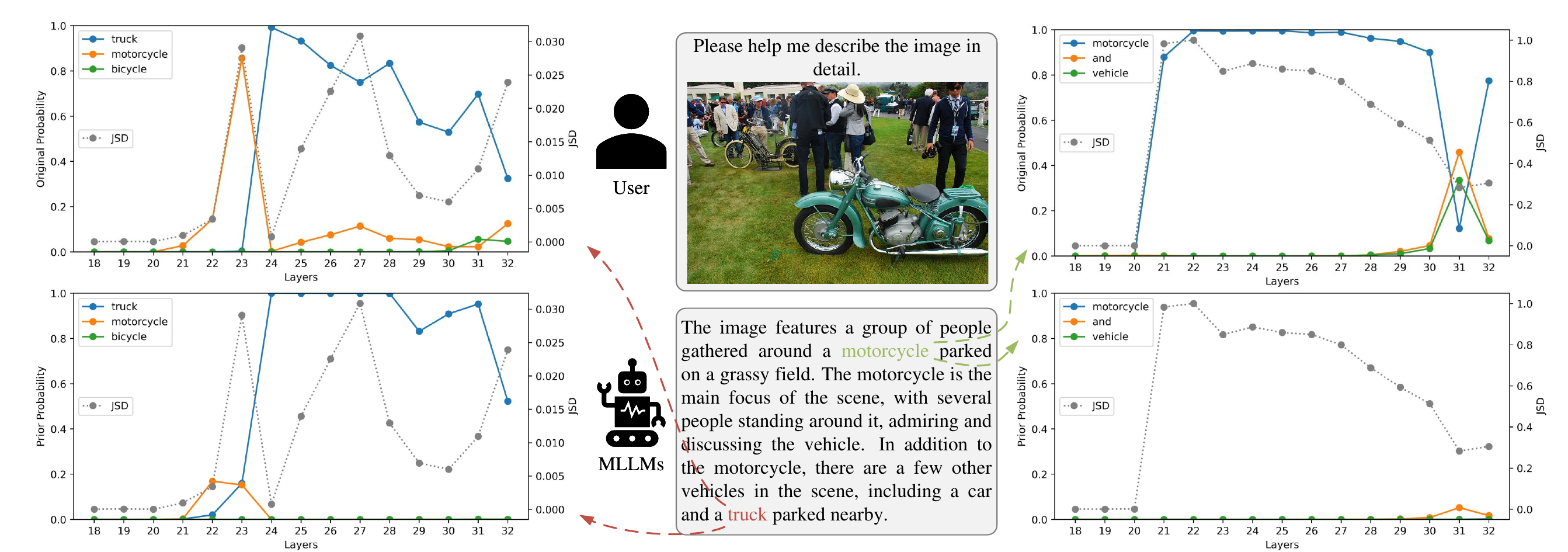}
\end{center}
   \caption{Illustration of the hierarchical evolution trends of hallucinated tokens (\textcolor{red}{red}), non-hallucinated tokens (\textcolor{teal}{green}), and JS divergence values across transformer layers, revealing that non-hallucinated tokens exhibit a trend similar to that of JS divergence values.}
\label{fig:2}
\end{figure}

Previous works~\cite{wang2024mllm, liang2024unleashing} have demonstrated that the representations in the preceding layers effectively capture (to some extent) visual information, while the deep-layer prior knowledge inherent to the Language Model (LLM) component systematically diminishes the likelihood of ground-truth tokens aligned with visual content. 
We leveraged the early-exit method~\cite{teerapittayanon2016branchynet, elbayad2019depth, schuster2022confident} to derive next-token probability distributions from intermediate layers for both original multimodal inputs (integrating visual and textual features) and text-only inputs (excluding visual components). Leveraging Jensen-Shannon (JS) divergence (detailed in Section~\ref{sec:3.3.1}), we quantify the dissimilarity between probability distributions of corresponding layers, as a proxy for the degree to which Language Model (LLM) prior knowledge suppresses visual information. This divergence is computed for each intermediate layer to characterize the hierarchical evolution of visual information suppression, from shallow to deep layers, induced by LLM priors.

Figure~\ref{fig:2} illustrates this hierarchical evolution with running examples. We analyze the Top-3 tokens ranked by probability in the final layer’s output. For the evolution trend in predicting the non-hallucinated token "\textit{motorcycle}", both the JS divergence value and the original probability of "\textit{motorcycle}" sharply increase at the 21\textsuperscript{st} 
layer, and then decrease at the 31\textsuperscript{st} layer. At the same time, the original probability and prior probability of the hallucinated token "\textit{and}" both increase at the 31\textsuperscript{st} layer, which indicates that as prior knowledge, "\textit{and}" suppresses the non-hallucinated token "\textit{motorcycle}" at the 31\textsuperscript{st} layer. For the evolutionary trajectory in predicting the hallucinated token "\textit{truck}", both the JS divergence value and the probability of the non-hallucinated token "\textit{motorcycle}" exhibit an upward trend starting at the 22\textsuperscript{nd} layer, reach a first peak at the 23\textsuperscript{rd} layer, followed by a simultaneous decline to minimum values at the 24\textsuperscript{th} layer, with consistent co-variation thereafter. Concurrently, the original probability and prior probability of the hallucinated token "\textit{truck}" increase from the 23\textsuperscript{rd} layer, peak at the 24\textsuperscript{th} layer, and then undergo gradual decay. This pattern suggests that the hallucinated token "\textit{truck}" also acts as prior knowledge to suppress the non-hallucinated token "\textit{motorcycle}".

Based on the above observations, we hypothesize that intermediate layers with higher JS divergence values contain richer visual factual knowledge, given the similar evolutionary trends between JS divergence values and non-hallucinated tokens. Therefore, we select the intermediate layer with the maximum JS divergence value as the target layer and then contrast its original probability distribution with the prior probability distribution to extract visual factual knowledge.

\subsection{Decoding by Extracting Visual Facts from Intermediate Layers}

Based on the above hypotheses, we propose Decoding by \textbf{E}xtracting \textbf{V}isual F\textbf{A}cts \textbf{(EVA)}, which can mitigate hallucinations during inference. As depicted in Figure~\ref{fig:1}, the EVA framework comprises two components: Dynamic Target Layer Selection (Section~\ref{sec:3.3.1}) and Decoding with Visual Factual Knowledge (Section~\ref{sec:3.3.2}).

\subsubsection{Dynamic Target Layer Selection}
\label{sec:3.3.1}

\textbf{Candidate Token Acquisition.} Due to the vast vocabulary space, we track only the changes in top-ranked tokens as candidate tokens across different layers to mitigate noise introduced by irrelevant tokens. This approach is grounded in the hypothesis that ground-truth tokens often appear in the top positions of an MLLM’s final-layer output logits~\cite{wang2024mllm}. We adopt the truncation strategy proposed in~\cite{wang2024mllm} to select candidate tokens, with the default strategy being top-p truncation. Formally:
\begin{equation}
    \mathcal{V}_{candidate}(x_T | x_{<T}) = \left\{ x_T \in \mathcal{V}: \sum_{v \in \mathcal{V}_p} P_{\tau}(x_T = v | x_0, x_1, \ldots, x_{T - 1}) \leq p \right\}
\end{equation}

where $\mathcal{V}$ is the whole vocabulary, and $p$ refers to the parameter used in top-$p$.

\textbf{Target Layer Selection.} We obtain the original probability distribution and prior probability distribution of the intermediate layers with early-exit method. Then, we calculate the JS divergence between the two distributions of the same layer. Formally:
\begin{equation} 
    d\left(p^j(x_T\mid x_{<T}), p^j_{prior}(x_T\mid x_{<T})\right) = \mathrm{JSD}\left(p^j(x_T\mid x_{<T}) || p^j_{prior}(x_T\mid x_{<T})\right), x_T \in \mathcal{V}_{candidate}
\end{equation}

where $\mathrm{JSD}(\cdot,\cdot)$ denotes the Jensen-Shannon divergence. Specifically, $p^j(x_T\mid x_{<T})$ represents the probability distribution over the next token derived from the original input at the $j$-th layer, while $p^j_{\text{prior}}(x_T\mid x_{<T})$ signifies the probability distribution over the next token derived from the pure text input at the $j$-th layer.  

Based on the observations in Section~\ref{sec:3.2}, we identify the target layer by selecting the one with the maximum JS divergence value for extracting visual factual knowledge. Formally, we have:

\begin{equation}
    M = \underset{j \in \mathcal{J}}{\arg\max} \{\mathrm{JSD}\left(p^j(x_T\mid x_{<T}) || p^j_{prior}(x_T\mid x_{<T})\right), x_T \in \mathcal{V}_{candidate}\}
\end{equation}

where $\mathcal{J} \subset \{0, \ldots, N - 1\}$ is a set of candidate layers.

\subsubsection{Decoding with Visual Factual Knowledge.} 
\label{sec:3.3.2}

\textbf{Dynamic Soft Modulation.} Inspired by~\cite{wang2024mllm}, we introduce two dynamic modulation coefficients, which are initialized as the maximum values of the original probability and the JS divergence from the selected target layer. Formally, we have:
\begin{align}
    &\text{max\_prob} = \text{max}(\text{softmax}(\phi(h_{T - 1}^M))). \\
    &\text{max\_JSD} = \mathrm{JSD}\left(p^M(x_T\mid x_{<T}) || p^M_{prior}(x_T\mid x_{<T})\right), x_T \in \mathcal{V}_{candidate}
\end{align}

\textbf{Visual Factual Knowledge Guided Decoding.} To extract visual factual knowledge, we subtract prior logits from the original logits at the target layer as visual factual knowledge. Subsequently, it is integrated into the original logits of the target layer for soft modulation, and the modulated logits are then proportionally merged into the final layer. We use a hyperparameter $\alpha$ to control the proportion of the extracted visual factual knowledge. By leveraging this knowledge for decoding, the probability of predicting the next token are updated as follows: 
\begin{align}
    &\Tilde{p}\left(x_T\mid x_{<T}\right) = \mathrm{softmax}(\mathrm{logits}) \\
    &\mathrm{logits} =\left(\phi\left(h_{T-1}^N\right)\right) + \alpha \times \text{max\_prob} \times (\phi\left(h_{T-1}^M\right)+\text{max\_JSD} \times \mathrm{logits}_v), \\
    &\mathrm{logits}_v = \phi\left(h_{T-1}^M\right)-\phi\left(\hat{h}_{T-1}^M\right)
\end{align}

where $N$ is the last layer of MLLM, $M$ is the selected preceding layer, $\phi\left(\hat{h}_{T-1}^M\right)$ is the prior logits.

\section{Experiments}
\label{sec:4}
\subsection{Experimental Settings}
\label{sec:4.1}
\textbf{Baselines.} We integrate EVA with diverse decoding strategies including greedy decoding, nucleus sampling, and beam search, and conduct comparative evaluations against multiple baselines for hallucination mitigation as detailed below: Dola~\cite{chuang2023dola} is specifically engineered to alleviate hallucinations in factual tasks for LLMs by diminishing shallow semantic interference to enhance the factuality of the final-layer output; VCD~\cite{leng2024mitigating} mitigates the dominance of language model priors in MLLMs through a subtractive mechanism that generates visual-information-enhanced representations by removing interfering prior knowledge at each sampling step; OPERA~\cite{huang2024opera} employs a dual strategy of dynamically penalizing overconfident tokens based on emerging aggregation patterns and introducing a retrospective allocation mechanism to preemptively address potential hallucinations that have already manifested; DeCo~\cite{wang2024mllm} dynamically selects preceding layers and leverages their prior knowledge to rectify final output logits for adjusting prediction biases. For all baselines, default hyperparameters from the original source code are strictly adopted to ensure fair comparability.

\textbf{Evaluation Models.} We select four of the most representative MLLM models for evaluation, including InstructBLIP~\cite{NEURIPS2023_9a6a435e}, MiniGPT-4~\cite{zhu2024minigpt}, LLaVA-1.5~\cite{liu2024improved} and Qwen-VL~\cite{bai2023qwen}. All evaluated MLLMs employ a language model size of 7 billion parameters (7B).

\textbf{Implementation Details.} For a 7B-parameter, 32-layer decoder-only architecture language model, we select layers 20–28 as candidates for the preceding layers, following~\cite{wang2024mllm}. For the image captioning and VQA tasks, the hyperparameter $\alpha$ is tuned within the range of 0.6 to 5. All experiments are conducted using a single L20 GPU for inference.

\subsection{Benchmark and metrics}

\textbf{POPE.} The Polling-based Object Probing Evaluation (POPE)~\cite{li2023evaluating} is a VQA-based metric designed to assess object hallucination in MLLMs. This framework evaluates hallucinations by posing questions of the form “Is there a <object> in the image?,” where <object> is drawn from three predefined splits: random, popular, and adversarial. The evaluation includes 500 MSCOCO images, with six questions per image for each split. We use the average F1-score computed on random, popular, and adversarial splits.

\textbf{MME.} The comprehensive MLLM Evaluation benchmark (MME)~\cite{fu2024mme} evaluates the perceptual and cognitive capabilities of MLLMs across 14 subtasks, covering domains such as OCR, visual knowledge, attribute relationships, and object recognition.

\textbf{CHAIR.} The Caption Hallucination Assessment with Image Relevance (CHAIR)~\cite{rohrbach2018object} metric, widely adopted in image captioning tasks, detects hallucinated objects by comparing extracted objects against ground-truth labels and evaluates hallucination at both the instance level ($\mathrm{CHAIR_I}$) and sentence level ($\mathrm{CHAIR_S}$), as formalized in Eq.~\ref{eq:10}. Following the experimental protocol in~\cite{huang2024opera}, we use consistent settings: 500 images from the MSCOCO 2014 validation dataset and the fixed prompt “Please help me describe the image in detail.” for evaluation.

\begin{equation}
\label{eq:10}
    \text{CHAIR}_\text{I} = \frac{|\{\text{hallucinated objects}\}|}{\text{all mentioned objects}}, \text{CHAIR}_\text{S} = \frac{|\{\text{captions with hallucinated objects}\}|}{\text{all captions}}.
\end{equation}

\subsection{Experimental Results}
\textbf{Results on POPE.} The experimental results on the POPE benchmark under random, popular, and adversarial settings are summarized in Table~\ref{tab:1}. A prominent observation is the consistent and robust performance of our proposed EVA method. Across different sampling scenarios, EVA consistently achieves improved F1-scores compared to baseline methods on all evaluated MLLMs, with the most notable improvement reaching 21.9 points under the nucleus decoding setting for MiniGPT-4. Moreover, EVA generally outperforms competing methods such as DoLa, DeCo, OPERA, and VCD, demonstrating its superior capability in suppressing hallucinations. EVA outperforms DeCo possibly because DeCo merely selects the layer with the maximum probability from intermediate layers, without considering that the maximum probability may be influenced by prior probability interference (as shown in Figure~\ref{fig:2}). This oversight leads to the selection of incorrect visual information for correcting the final output. In conclusion, the performance gain highlights EVA’s effectiveness in mitigating object hallucinations by extracting visual factual knowledge.

\begin{table}[h]
\centering
\caption{POPE hallucination evaluation results. The best results are in bold.}
\label{tab:1}
\resizebox{0.7\textwidth}{!}{
\begin{tabular}{@{}llllll@{}}
\toprule
\multirow{2}{*}{Decoding} & \multirow{2}{*}{Method} & \multicolumn{1}{c}{InstructBLIP} & \multicolumn{1}{c}{MiniGPT-4} & \multicolumn{1}{c}{LLaVA-1.5} & \multicolumn{1}{c}{Qwen-VL} \\
&  & F1 $\uparrow$ & F1 $\uparrow$ & F1 $\uparrow$ & F1 $\uparrow$ \\ \midrule
\multirow{4}{*}{Greedy} 
& Vanilla & 80.0 & 58.5 & 82.2 & 85.2 \\
& DoLa & 83.4 & 72.8 & 83.2 & 85.8 \\
& DeCo & 84.9 & 77.4 & 86.7 & 86.3 \\
& \textbf{EVA (Ours)} & \textbf{85.2 \textcolor{blue}{$\uparrow$5.2}} & \textbf{78.2 \textcolor{blue}{$\uparrow$19.7}} & \textbf{87.1 \textcolor{blue}{$\uparrow$4.9}} & \textbf{86.3 \textcolor{blue}{$\uparrow$1.1}} \\ \midrule
\multirow{4}{*}{Beam Search} 
& Vanilla & 84.4 & 70.3 & 84.9 & 85.3 \\
& OPERA & 84.8 & 73.3 & 85.4 & 86.1 \\
& DeCo & 84.9 & 77.9 & 86.7 & \textbf{86.4} \\
& \textbf{EVA (Ours)} & \textbf{85.0 \textcolor{blue}{$\uparrow$0.6}} & \textbf{78.0 \textcolor{blue}{$\uparrow$7.7}} & \textbf{87.1 \textcolor{blue}{$\uparrow$2.2}} & 86.2 \textbf{\textcolor{blue}{$\uparrow$0.9}} \\ \midrule
\multirow{4}{*}{Nucleus} 
& Vanilla & 79.8 & 52.8 & 83.1 & 84.5 \\
& VCD & 79.9 & 56.0 & 83.1 & 84.7 \\
& DeCo & 81.8 & 63.8 & 85.4 & 85.2 \\
& \textbf{EVA (Ours)} & \textbf{84.5 \textcolor{blue}{$\uparrow$4.7}} & \textbf{74.7 \textcolor{blue}{$\uparrow$21.9}} & \textbf{87.4 \textcolor{blue}{$\uparrow$4.3}} & \textbf{85.4 \textcolor{blue}{$\uparrow$0.9}} \\ \bottomrule
\end{tabular}
}
\end{table}

\textbf{Results on MME Hallucination Subset.} The evaluation on the MME subset extends beyond the scope of POPE, encompassing hallucination issues at both object and attribute levels. Results in Table~\ref{tab:2} demonstrate that the EVA method yields overall positive effects on object-level metrics, contributing to significant improvements in overall performance. Although EVA’s total score on the InstructBLIP model does not achieve the highest value (slightly below VCD’s 447.67), it attains the highest total scores on other models, reflecting its effectiveness and stability in enhancing hallucination mitigation across MLLMs. These improvements highlight the importance of EVA’s effective utilization of visual information, thereby exerting a positive impact on a broader range of hallucination challenges.

\begin{table}[h]
\centering
\caption{Results on the hallucination subset of MME with nucleus sampling. The best results within each setting are bolded.}
\label{tab:2}
\resizebox{\textwidth}{!}{
\begin{tabular}{@{}clllllll@{}}
\toprule
\multirow{2}{*}{Model} & \multirow{2}{*}{Decoding} & \multicolumn{2}{c}{Object-level} & \multicolumn{2}{c}{Attribute-level} & \multirow{2}{*}{Total Scores$\uparrow$} \\ \cmidrule(lr){3 - 4} \cmidrule(lr){5 - 6}  
&  & Existence$\uparrow$ & Count$\uparrow$ & Position$\uparrow$ & Color$\uparrow$ &  \\ \midrule
\multirow{4}{*}{LLaVA-1.5}  
& Nucleus & 175.67 & 124.67 & 114.00 & 151.00 & 565.33 \\ 
& VCD & 184.66 & 138.33 & 128.67 & 153.00 & 604.66 \\  
& DeCo & 190.00 & 148.33 & 115.00 & 165.00 & 618.33 \\  
& Ours & \textbf{190.00} \textbf{\textcolor{blue}{$\uparrow$14.33}} & \textbf{158.33} \textbf{\textcolor{blue}{$\uparrow$33.66}} & \textbf{133.33} \textbf{\textcolor{blue}{$\uparrow$19.33}} & \textbf{170.00} \textbf{\textcolor{blue}{$\uparrow$19.00}} & \textbf{651.67} \textbf{\textcolor{blue}{$\uparrow$86.34}} \\ 
\midrule
\multirow{4}{*}{Qwen-VL}  
& Nucleus & 155.00 & 127.67 & \textbf{131.67} & 173.00 & 587.33 \\ 
& VCD & 156.00 & 131.00 & 128.00 & 181.67 & 596.67 \\  
& DeCo & \textbf{185.00} & 135.00 & 106.67 & 175.00 & 601.67 \\  
& Ours & 175.00 \textbf{\textcolor{blue}{$\uparrow$20.00}} & \textbf{135.00} \textbf{\textcolor{blue}{$\uparrow$7.33}} & 115.00 \textbf{\textcolor{purple}{$\downarrow$16.67}} & \textbf{185.00} \textbf{\textcolor{blue}{$\uparrow$12.00}} & \textbf{610.00} \textbf{\textcolor{blue}{$\uparrow$22.67}} \\ 
\midrule
\multirow{4}{*}{InstructBLIP}  
& Nucleus & 141.00 & 75.33 & \textbf{66.67} & 97.33 & 380.33 \\ 
& VCD & 168.33 & \textbf{92.33} & 64.00 & \textbf{123.00} & \textbf{447.67} \\  
& DeCo & 185.00 & 55.00 & 48.33 & 100.00 & 388.33 \\  
& Ours & \textbf{185.00} \textbf{\textcolor{blue}{$\uparrow$44.00}} & 60.00 \textbf{\textcolor{purple}{$\downarrow$15.33}} & 55.00 \textbf{\textcolor{purple}{$\downarrow$11.67}} & 115.00 \textbf{\textcolor{blue}{$\uparrow$17.67}} & 415.00 \textbf{\textcolor{blue}{$\uparrow$34.67}}\\ 
\midrule
\multirow{4}{*}{MiniGPT-4}
& Nucleus & 65.00 & 70.00 & 46.67 & 56.67 & 238.33 \\ 
& VCD & \textbf{105.00} & 61.67 & 68.33 & \textbf{85.00} & 320.00 \\  
& DeCo & 75.00 & 70.00 & 53.33 & 65.00 & 263.33 \\  
& Ours & 103.33 \textbf{\textcolor{blue}{$\uparrow$38.33}} & \textbf{78.33} \textbf{\textcolor{blue}{$\uparrow$8.33}} & \textbf{75.00} \textbf{\textcolor{blue}{$\uparrow$28.33}} & 65.00 \textbf{\textcolor{blue}{$\uparrow$8.33}} & \textbf{321.67} \textbf{\textcolor{blue}{$\uparrow$83.34}}\\ 
\bottomrule
\end{tabular}
}
\end{table}

\begin{figure}[htbp]
\begin{center}
   \includegraphics[width=\linewidth]{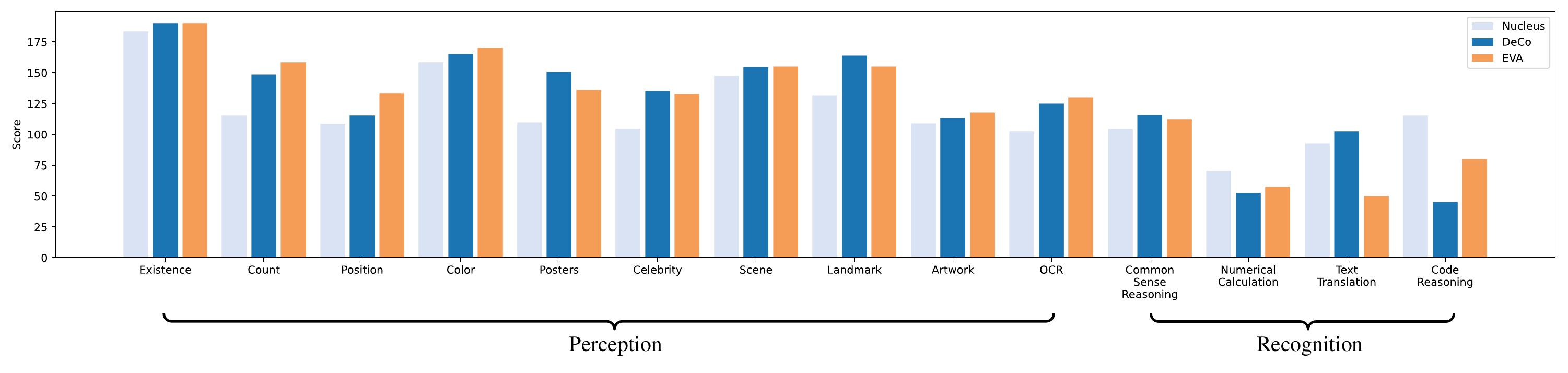}
\end{center}
   \caption{
   MME full set results on LLaVA-1.5 with nucleus sampling. EVA leads to substantial enhancement in LVLMs’ perception capacities.}
\label{fig:3}
\end{figure}

\textbf{Results on MME Hallucination Full Set.} As illustrated in Figure~\ref{fig:3}, we also evaluate EVA on the MME Full Set to assess its impact on the overall capabilities of MLLMs. Given the comparable performance trends across all models, we present results for LLaVA-1.5 as a representative case. The application of EVA leads to substantial improvements in perception-based tasks. However, this is accompanied by a reduction in the original recognition capabilities of MLLMs, which may be attributed to EVA’s mechanism of subtracting LLM priors, thereby causing a decline in recognition capabilities.

\textbf{Results on CHAIR.}  From Table~\ref{tab:3}, we observe that EVA predominantly outperforms other approaches in mitigating hallucinations across four MLLMs. The proposed EVA approach effectively reduces hallucinations in visual description tasks through the extraction of visual factual knowledge.


\begin{table}[h]
\centering
\caption{CHAIR hallucination evaluation results. Lower scores indicate fewer hallucinations. The best results are in bold. *Qwen-VL may generate relatively short sentences leading to excessively low results. Note that we use the baseline’s original decoding settings for a fair comparison and run EVA under the same settings.}
\label{tab:3}
\resizebox{\textwidth}{!}{
\begin{tabular}{@{}lllllllllllll@{}}
\toprule
\multirow{2}{*}{Decoding} & \multirow{2}{*}{Method} & \multicolumn{2}{c}{InstructBLIP} & \multicolumn{2}{c}{MiniGPT-4} & \multicolumn{2}{c}{LLaVA-1.5} & \multicolumn{2}{c}{Qwen-VL} \\ \cmidrule(lr){3 - 4} \cmidrule(lr){5 - 6} \cmidrule(lr){7 - 8} \cmidrule(lr){9 - 10} 
 &  & $\text{CHAIR}_\text{S} \downarrow$ & $\text{CHAIR}_\text{I} \downarrow$ & $\text{CHAIR}_\text{S} \downarrow$ & $\text{CHAIR}_\text{I} \downarrow$ & $\text{CHAIR}_\text{S} \downarrow$ & $\text{CHAIR}_\text{I} \downarrow$ & $\text{CHAIR}_\text{S} \downarrow$ & $\text{CHAIR}_\text{I} \downarrow$ \\ \midrule
\multirow{4}{*}{Greedy} 
 & Vanilla & 58.8 & 23.7 & 31.8 & 9.9 & 45.0 & 14.7 & 46.0 & 12.5 \\
 & DoLa & 48.4 & 15.9 & 32.2 & 10.0 & 47.8 & 13.8 & 46.8 & 12.9 \\
 & DeCo & 41.2 & 14.4 & 27.0 & \textbf{8.8} & 37.8 & \textbf{11.1} & 42.2 & 10.7 \\
 & \textbf{EVA (Ours)} & \textbf{41.2 \textcolor{blue}{$\downarrow$17.6}} & \textbf{13.0 \textcolor{blue}{$\downarrow$10.7}} & \textbf{26.0 \textcolor{blue}{$\downarrow$5.8}} & 9.2 \textbf{\textcolor{blue}{$\downarrow$0.7}} & \textbf{35.6 \textcolor{blue}{$\downarrow$9.4}} & 11.3 \textbf{\textcolor{blue}{$\downarrow$3.4}} & \textbf{6.8 \textcolor{blue}{$\downarrow$39.2}} & \textbf{4.4 \textcolor{blue}{$\downarrow$8.1}} \\ \midrule 
\multirow{4}{*}{Beam Search} 
 & Vanilla & 55.6 & 15.8 & 30.6 & 9.5 & 48.8 & 13.9 & 41.8 & 10.8 \\
 & OPERA & 46.4 & 14.2 & 26.2 & 9.5 & 44.6 & 12.8 & 34.6 & 9.5 \\
 & DeCo & 43.8 & 12.7 & 24.8 & \textbf{7.5} & \textbf{33.0} & \textbf{9.7} & 32.0 & 8.7 \\
 & \textbf{EVA (Ours)} & \textbf{40.6 \textcolor{blue}{$\downarrow$15.0}} & \textbf{12.2 \textcolor{blue}{$\downarrow$3.6}} & \textbf{22.6 \textcolor{blue}{$\downarrow$8.0}} & 8.0 \textbf{\textcolor{blue}{$\downarrow$1.5}} & 37.4 \textbf{\textcolor{blue}{$\downarrow$11.4}} & 11.0 \textbf{\textcolor{blue}{$\downarrow$2.9}} & \textbf{2.8 \textcolor{blue}{$\downarrow$39.0}} & \textbf{2.0 \textcolor{blue}{$\downarrow$8.8}} \\ \midrule 
\multirow{4}{*}{Nucleus} 
 & Vanilla & 54.6 & 24.8 & 32.6 & 10.7 & 48.8 & 14.2 & 49.2 & 13.1 \\
 & VCD & 58.0 & 17.0 & 33.8 & 11.1 & 54.0 & 16.0 & 46.4 & 11.9 \\
 & DeCo & 43.6 & \textbf{12.9} & 30.8 & \textbf{9.5} & 42.8 & 13.2 & 43.8 & 11.8 \\
 & \textbf{EVA (Ours)} & \textbf{39.0 \textcolor{blue}{$\downarrow$15.6}} & 15.0 \textbf{\textcolor{blue}{$\downarrow$9.8}} & \textbf{26.0 \textcolor{blue}{$\downarrow$6.6}} & 10.1 \textbf{\textcolor{blue}{$\downarrow$0.6}} & \textbf{40.0 \textcolor{blue}{$\downarrow$8.8}} & \textbf{11.4 \textcolor{blue}{$\downarrow$2.8}} & \textbf{7.4 \textcolor{blue}{$\downarrow$41.8}} & \textbf{4.1 \textcolor{blue}{$\downarrow$9.0}} \\ \bottomrule
\end{tabular}
}
\end{table}

\subsection{Discussions}
\textbf{Effectiveness of Visual Knowledge Extraction.} To validate the effectiveness of our approach, we conduct an experiment comparing our visual knowledge extraction method with DeCo~\cite{wang2024mllm}. Following~\cite{wang2024mllm}, we randomly sample 500 images from the MSCOCO dataset and use prompts to elicit raw responses from LLaVA1.5-7B. We then extract all non-existent objects mentioned in the responses, along with their preceding contextual text, and feed this data into the MLLM.  We observe the next-token probabilities at the selected layer to derive candidate tokens, denoted as $\mathcal{V}_{candidate}$, with a default confidence threshold of 0.9. We then label the tokens in $\mathcal{V}_{candidate}$. Specifically, we filter out data where $\mathcal{V}_{candidate}$ contains at least one ground truth token and observe whether an activated ground truth token exists among the candidate tokens, formally expressed as:

\begin{equation}
    \exists x_a \in \mathcal{V}_{candidate}, \ p(x_a | x_{<s}) - p(x_h | x_{<s}) > 0
\end{equation}

where $x_a$ denotes the activated ground-truth token, and $x_h$ represents the token with the highest probability of being a hallucinated token in the probability distribution of the final layer. Additionally, we count the total number of activated ground truth token. Based on the experimental setup described above, we compare the statistical results derived from the following two probabilities:
\begin{align}
    &p_v(\cdot \mid x_{<s})=\mathrm{softmax}(\phi\left(h_{T-1}^M\right)-\phi\left(\hat{h}_{T-1}^M\right)), \\
    &p_d(\cdot \mid x_{<s})=\mathrm{softmax}(\phi\left(h_{T-1}^\mathcal{A}\right)), \\
    &\mathcal{A} = \underset{j \in \mathcal{J}}{\arg\max} \{p^j(x_T\mid x_{<T}), x_T \in \mathcal{V}_{candidate}\}
\end{align}

where $h_{T-1}^\mathcal{A}$ is the the hidden state of the target layer selected by DeCo~\cite{wang2024mllm} method, $p_v(\cdot \mid x_{<s})$ represents the probability distribution of visual knowledge extracted by our method, while $p_d(\cdot \mid x_{<s})$ denotes that of DeCo\cite{wang2024mllm}.

\begin{table}[htbp]
\centering
\caption{Statistical results obtained from ours and DeCo~\cite{wang2024mllm}.}
\label{tab:4}
\begin{tabular}{ccc}
\hline
 Method & Number of Data & Number of Token \\ \hline
 DeCo & 157 & 259 \\ \hline
 Ours & \textbf{170} & \textbf{302} \\ \hline
\end{tabular}
\end{table}

As shown in Table~\ref{tab:4}, the statistical results demonstrate that our method yields a larger quantity of data containing at least one ground truth token and a higher count of activated ground truth tokens. This suggests that our method is more effective in extracting visual factual knowledge.

\begin{figure}[htbp]
\begin{center}
   \includegraphics[width=\linewidth]{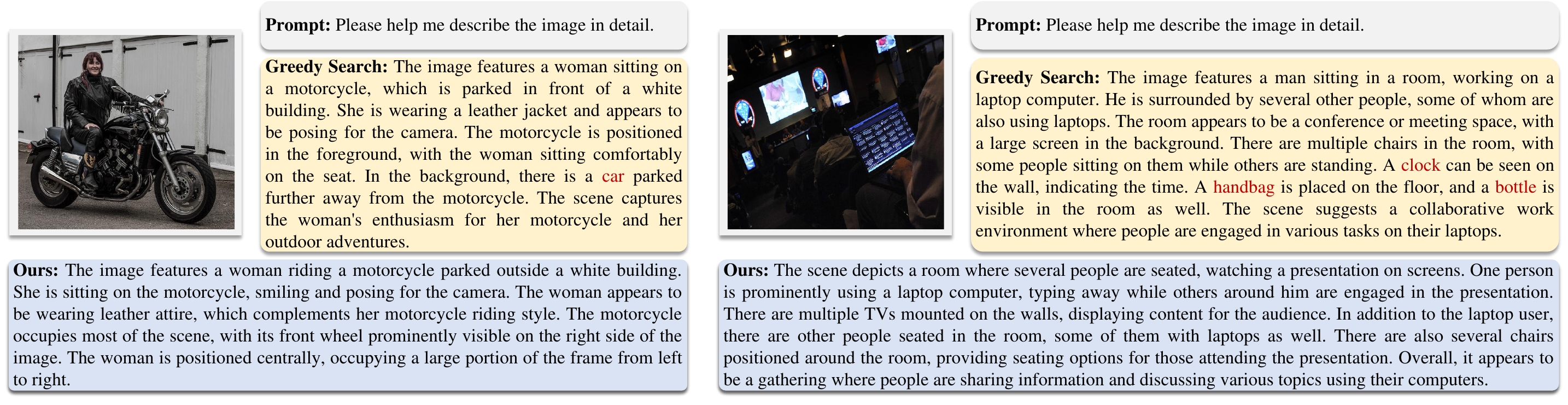}
\end{center}
   \caption{
   Illustration of hallucination correction by EVA with two samples from LLaVA-Bench. Hallucinated objects from LVLM’s greedy search are highlighted in \textcolor{red}{red}.}
\label{fig:4}
\end{figure}

\textbf{Case Study on LLaVA-Bench.} Figure~\ref{fig:4} presents two illustrative case studies demonstrating that, given identical prompts and images, conventional decoding methods often produce object hallucinations driven by inherent statistical biases and language priors acquired during pretraining. Specifically, the examples reveal hallucinated objects such as “clock,” “handbag,” and “bottle,” which frequently co-occur with the true context object “room.” In contrast, our proposed EVA method effectively mitigates these hallucination artifacts while maintaining the coherence and informativeness of the generated text. This highlights EVA’s ability to dynamically incorporate visual factual knowledge, resulting in more accurate and reliable multimodal outputs

\section{Conclusion and Limitation}
\label{sec:5}

In this paper, we investigate how prior knowledge in MLLMs suppresses visual factual knowledge at the intermediate layer stage. We find that visual factual knowledge exhibits an evolutionary trend analogous to the divergence between raw probability distributions and prior probability distributions in intermediate layers. Based on this finding, we propose EVA, a train-free decoding method that revises final outputs by extracting visual factual knowledge from intermediate layers. We validated EVA on widely used benchmarks, and the results show that compared with baseline methods, EVA significantly reduces the incidence of hallucinations, highlighting its effectiveness in mitigating hallucinations.

\textbf{Limitation.} Due to GPU cost constraints, our experiments are exclusively conducted on a limited set of MLLMs, without exploring additional models or those with larger parameter scales.

\newpage

{
\bibliographystyle{IEEEtran}
\bibliography{main}
}


\newpage
\appendix

\large{\textbf{Appendix}}

\section{Ablation Results of Dynamic Soft Modulation}

To quantify the effect of soft modulation, we evaluated the results of removing the "\text{max\_JSD}", removing the "\text{max\_prob}", and removing both with greedy search. The ablation results are shown in Table~\ref{tab:8}.

\begin{table}[h]
\centering
\caption{Ablation study of soft modulation.}
\label{tab:8}
\resizebox{0.7\textwidth}{!}{
\begin{tabular}{@{}lll@{}}
\toprule
\multirow{2}{*}{Method} & \multicolumn{1}{c}{MiniGPT-4} & \multicolumn{1}{c}{Qwen-VL} \\
& F1 $\uparrow$ & F1 $\uparrow$ \\ \midrule
EVA & \textbf{78.2} & \textbf{86.3} \\
EVA(w/o \text{max\_JSD}) & 77.8 & 86.1 \\
EVA(w/o \text{max\_prob}) & 76.9 & 84.6 \\
EVA(w/o \text{max\_JSD} \& \text{max\_prob}) & 76.6 & 84.5 \\ \bottomrule
\end{tabular}
}
\end{table}

\section{Case Studies across Diverse MLLMs}

\begin{figure}[htbp]
\begin{center}
   \includegraphics[width=\linewidth]{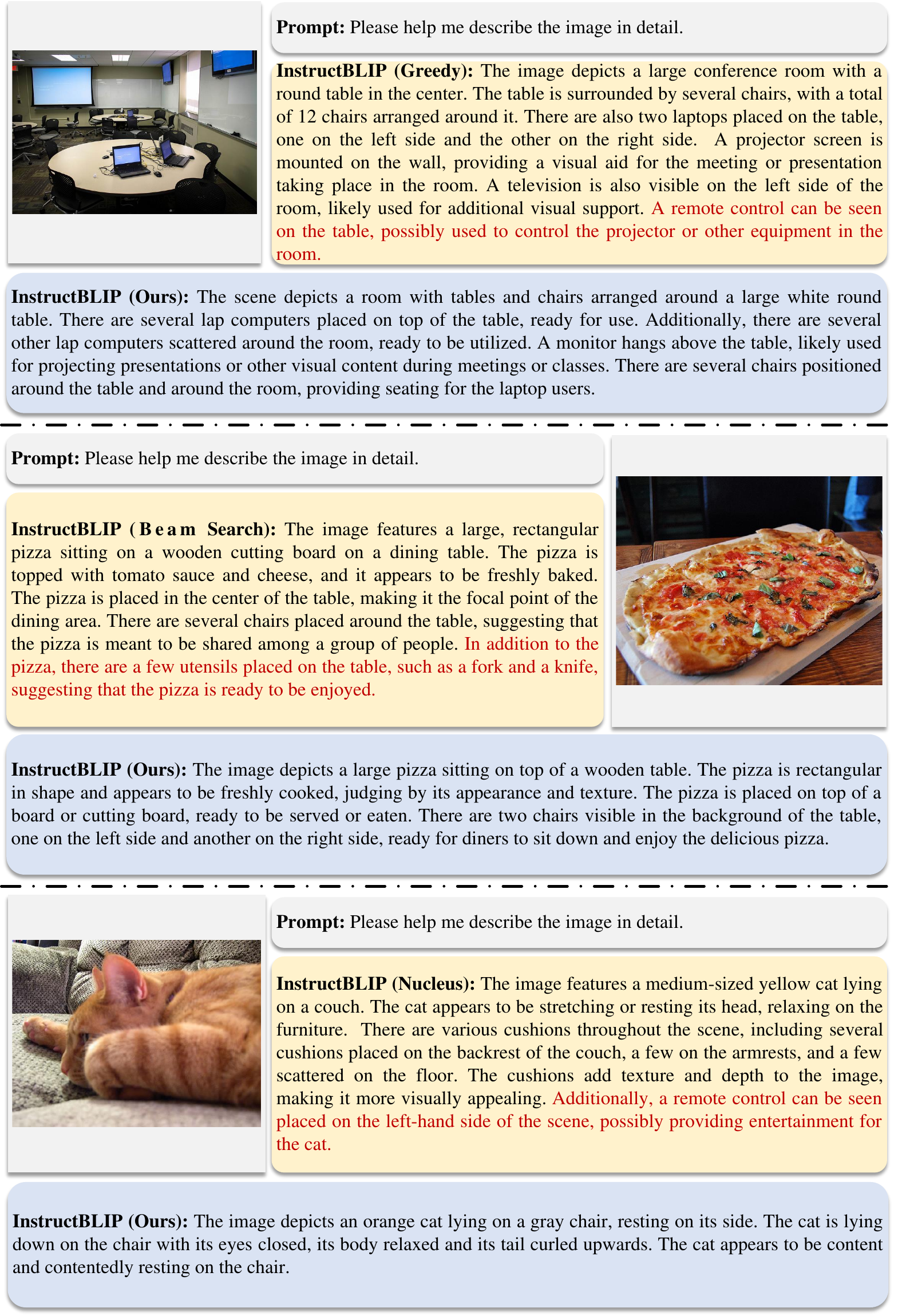}
\end{center}
   \caption{EVA’s performance in mitigating hallucinations of InstructBLIP on three decoing methods. Hallucinated content is highlighted in \textcolor{red}{red}.}
\label{fig:5}
\end{figure}

\begin{figure}[htbp]
\begin{center}
   \includegraphics[width=\linewidth]{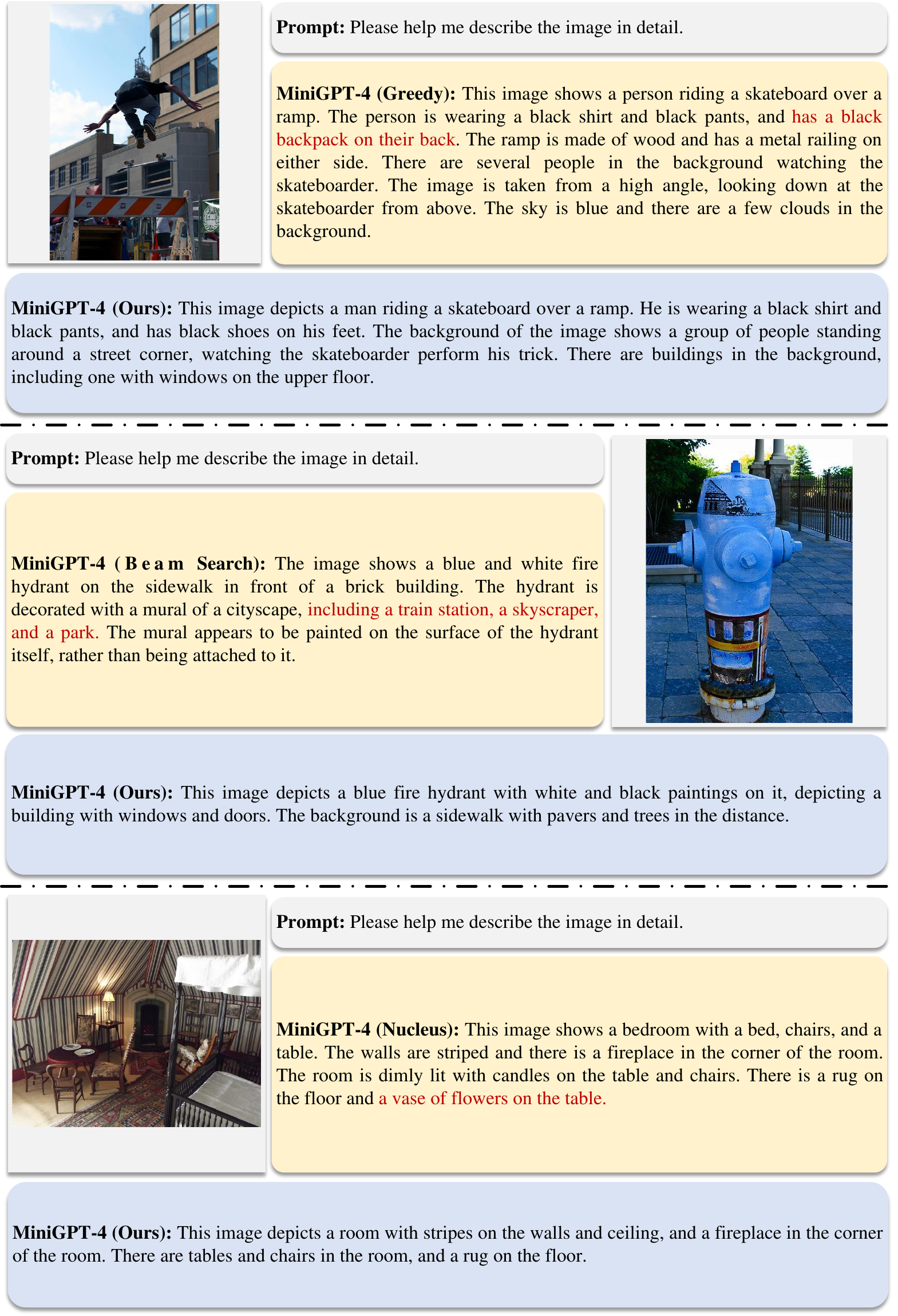}
\end{center}
   \caption{EVA’s performance in mitigating hallucinations of MiniGPT-4 on three decoing methods. Hallucinated content is highlighted in \textcolor{red}{red}.}
\label{fig:6}
\end{figure}

\begin{figure}[htbp]
\begin{center}
   \includegraphics[width=\linewidth]{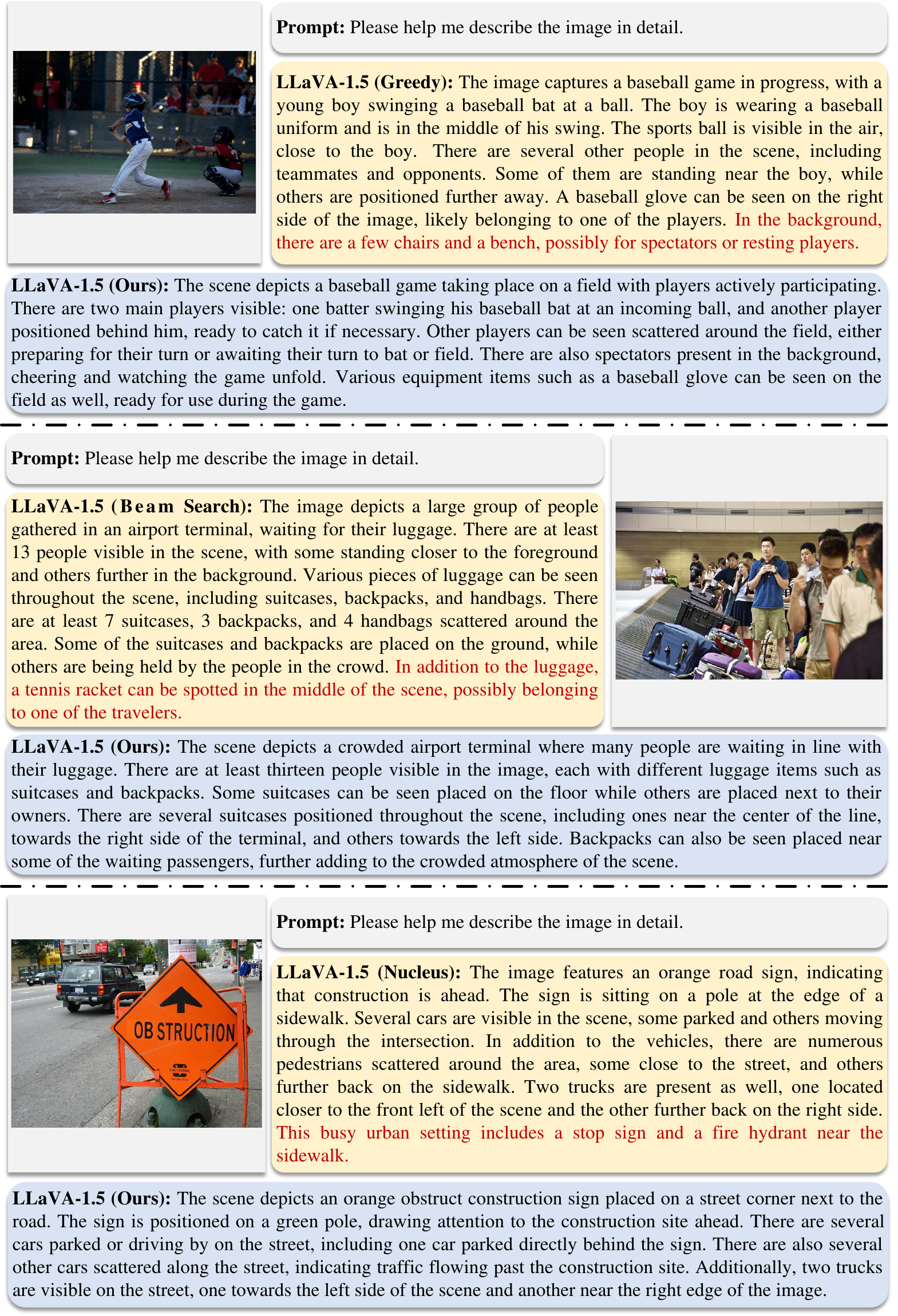}
\end{center}
   \caption{EVA’s performance in mitigating hallucinations of LLaVA-1.5 on three decoing methods. Hallucinated content is highlighted in \textcolor{red}{red}.}
\label{fig:7}
\end{figure}

\begin{figure}[htbp]
\begin{center}
   \includegraphics[width=\linewidth]{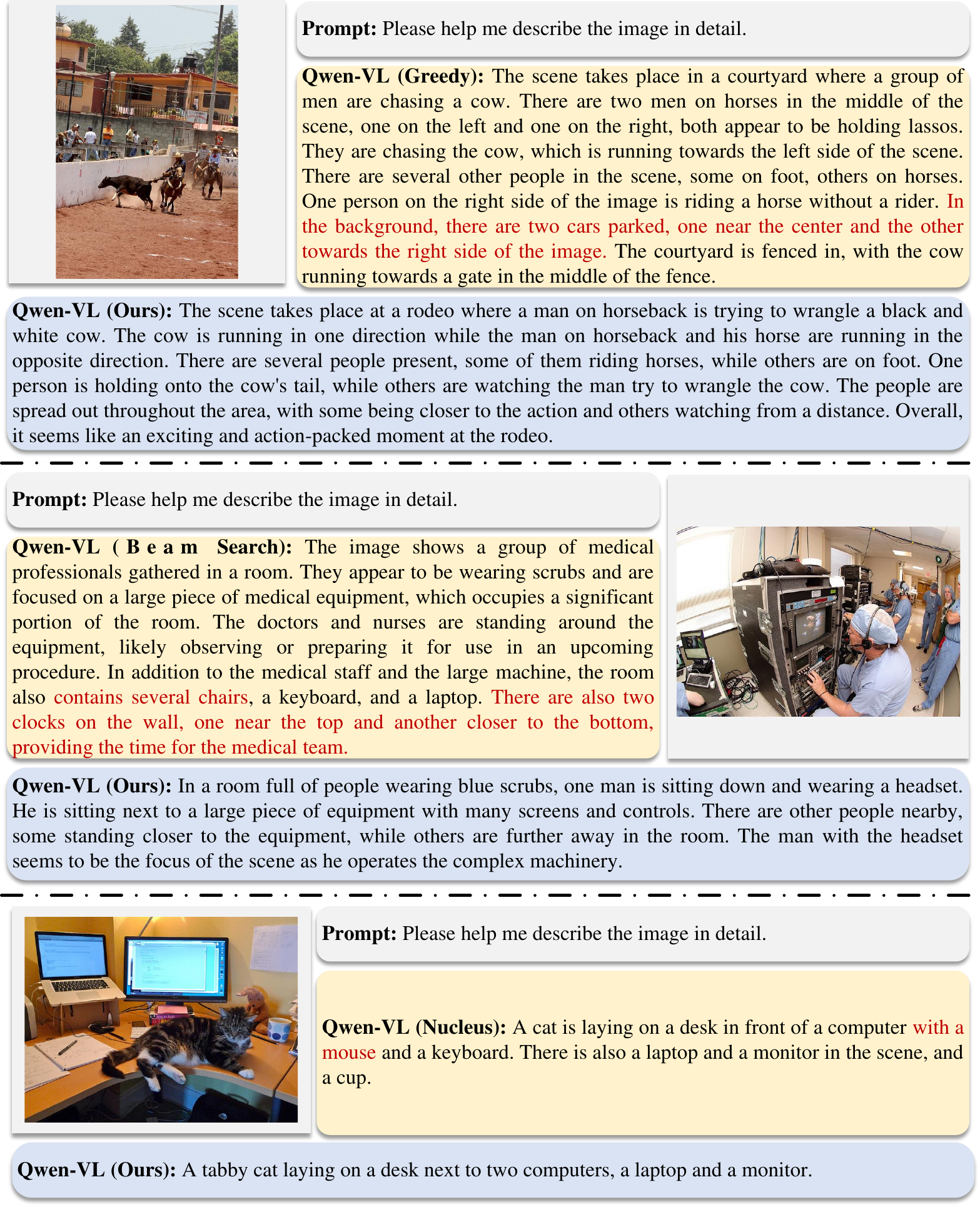}
\end{center}
   \caption{EVA’s performance in mitigating hallucinations of Qwen-VL on three decoing methods. Hallucinated content is highlighted in \textcolor{red}{red}.}
\label{fig:8}
\end{figure}

Additional case studies across diverse MLLMs are provided to demonstrate the effectiveness of EVA. Specifically, Figures~\ref{fig:5}, \ref{fig:6}, \ref{fig:7}, and \ref{fig:8} respectively present further instances of hallucination corrections by our method for InstructBLIP, MiniGPT-4, LLaVA-1.5, and Qwen-VL.  

\end{document}